\pdfoutput=1

\documentclass[11pt]{article}

\usepackage{acl}

\usepackage{times}
\usepackage{latexsym}

\usepackage[T1]{fontenc}

\usepackage[utf8]{inputenc}

\usepackage{microtype}

\usepackage{inconsolata}

\usepackage{graphicx}
\usepackage{subfigure}
\usepackage{subcaption}
\usepackage{caption}
\usepackage{xcolor}
\usepackage{multirow}
\usepackage{cleveref}
\usepackage{array}
\usepackage{rotating}
\usepackage{booktabs}

\title{Enhancing Few-Shot Stock Trend Prediction with Large Language Models}

\author{Yiqi Deng\textsuperscript{1}\thanks{Work in progress.}, Xingwei He\textsuperscript{1}, Jiahao Hu\textsuperscript{1}, Siu-Ming Yiu\textsuperscript{1}\\
  The University of Hong Kong \\ 
\texttt{yqdeng@cs.hku.hk, hexingwei15@gmail.com} \\
\texttt{hujh0919@gmail.com, smyiu@cs.hku.hk} \\
}

\begin{document}
\maketitle
\begin{abstract}
The goal of stock trend prediction is to forecast future market movements for informed investment decisions. Existing methods mostly focus on predicting stock trends with supervised models trained on extensive annotated data.  However, human annotation can be resource-intensive and the annotated data are not readily available. Inspired by the impressive few-shot capability of Large Language Models (LLMs), we propose using LLMs in a few-shot setting to overcome the scarcity of labeled data and make prediction more feasible to investors. 
Previous works typically merge multiple financial news for predicting stock trends, causing two significant problems when using LLMs: (1) Merged news contains noise, and (2) it may exceed LLMs' input limits,
leading to performance degradation.  
To overcome these issues, we propose a two-step method `\textit{denoising-then-voting}'. Specifically, we introduce an `Irrelevant' category, and predict stock trends for individual news instead of merged news. Then we aggregate these predictions using majority voting. The proposed method offers two advantages: (1) Classifying noisy news as irrelevant removes its impact on the final prediction. (2) Predicting for individual news mitigates LLMs' input length limits.
Our method achieves 66.59\% accuracy in S\&P 500, 62.17\% in CSI-100, and 61.17\% in HK stock prediction, outperforming the standard few-shot counterparts by around 7\%, 4\%, and 4\%. Furthermore, our proposed method performs on par with state-of-the-art supervised methods.
\end{abstract}

\section{Introduction}

Stock trend prediction aims to forecast future movements as upward or downward for specific stocks or indexes. Accurate projections on stocks benefit investors or consultants in their trading activities like designing profitable trading strategies~\cite{sawhney2021fast,sawhney-etal-2021-quantitative}, portfolio optimization~\cite{du2020stock,niu2022metatrader}, and risk management~\cite{zhu2022wise}. 
Typically, stock trend prediction entails the analysis of historical market data, such as price and volume. 
Financial news also plays an important role in influencing how markets move~\cite{tetlock2007giving,ashtiani2023news},
for instance, economic reports reflecting the financial health of one company, announcements of mergers and acquisitions, and declarations of leadership changes. 
In addition, by analyzing the sentiment, tone, and keywords in news texts, it is possible to identify patterns that may indicate future stock price movements.

\begin{figure}[!t]
\centering
\includegraphics[width=0.45\textwidth]{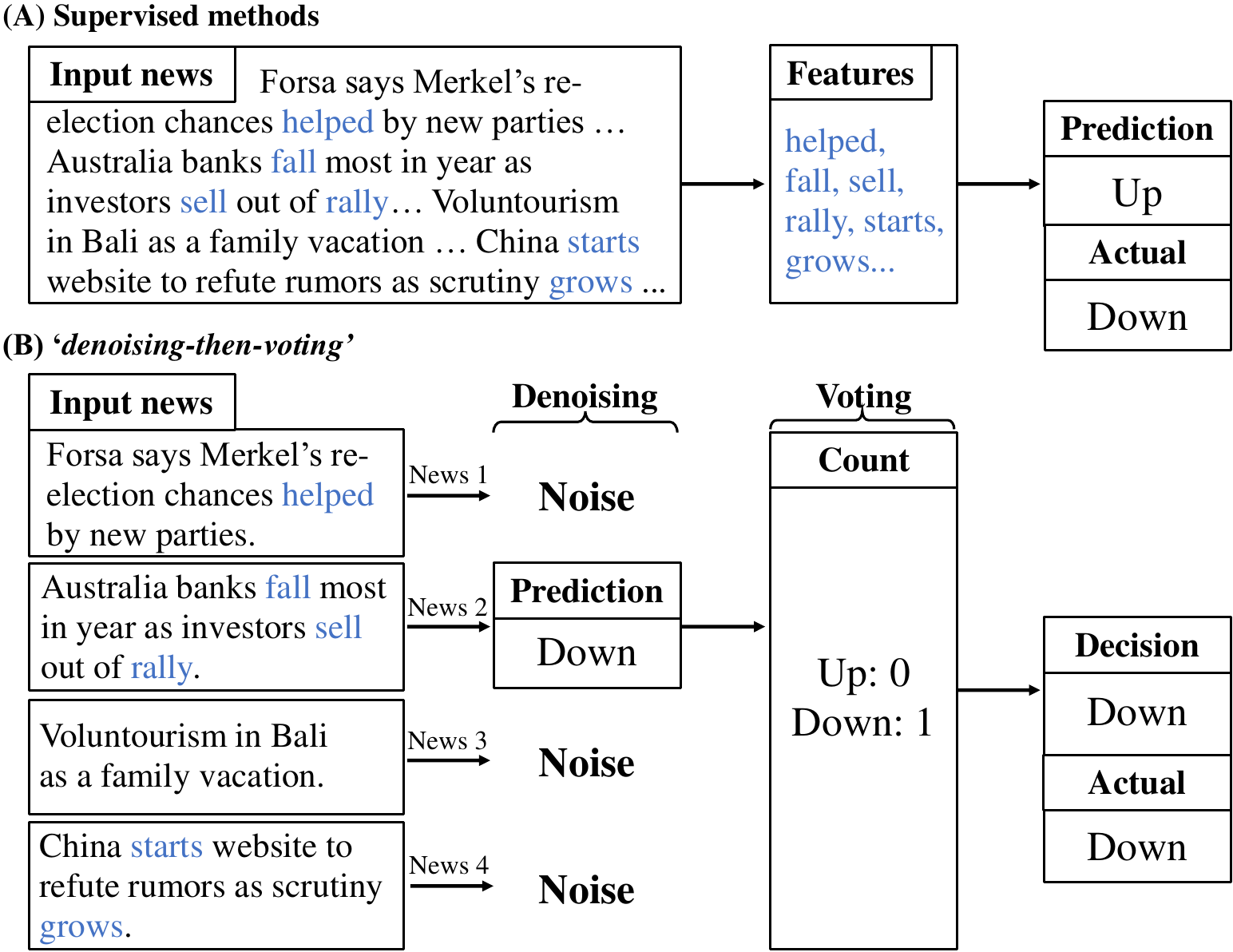} \caption{Comparison between state-of-the-art supervised methods and our proposed LLM-based `\textit{denoising-then-voting}'method.}
\label{fig:Intro}
\end{figure}

To extract information from financial news for stock trend prediction, supervised methods have emerged as a prevalent choice ~\cite{yang2018explainable,chen2019investment,li2021modeling}.  
However, 
supervised methods heavily rely on labeled data. Unfortunately, 
it is challenging to obtain labeled data due to the laborious and time-consuming annotation process, which hinders the practical applicability of supervised models. 
Recent works~\cite{wei2022chain,he2023annollm} have highlighted the remarkable few-shot ability of large language models (LLMs) in various downstream NLP tasks. 
Motivated by this, we propose harnessing the power of LLMs, especially ChatGPT, 
to alleviate the need for labeled financial data and 
make prediction more feasible for investors in real-world financial practices.

To adopt LLMs in stock prediction, two problems arise if we follow the previous work~\cite{huang2022news,DENG2023121654} to merge multiple news as inputs and directly predict stock trends.
First, the merged news contains noise, which may cause incorrect predictions. 
In Figure~\ref{fig:Intro}(A), features (`\textit{helped}', `\textit{starts}', `\textit{grows}') from news (News \#1, \#4) that is poorly related to the trend can lead to an incorrect prediction of `Up' trend.
Previous supervised works~\cite{ding-etal-2019-event-representation,huang2018tensor} can effectively capture features from news, but it is challenging to discern whether news being analyzed is relevant to trends or not. 
Second, the merged news may exceed the maximum input length of LLMs\footnote{An upper input limit for GPT-3.5 is 4,097 tokens. Please refer to \url{https://platform.openai.com/docs/models}.}. 
Simply truncating or shortening the merged news can result in a loss of crucial information about stock trends, leading to a degraded performance in the final prediction.
To address these issues, we propose a two-step method `\textit{denoising-then-voting}' as shown in Figure~\ref{fig:Intro}(B). 
We first identify and filter out the irrelevant news through a denoising strategy.
Then we aggregate all relevant news and predict stock trends via majority voting.
The proposed `\textit{denoising-then-voting}' method allows us to focus on denoised news without losses of useful information relevant to the stock trend, which helps to improve prediction accuracy.

To summarize, our contributions are as follows: 
(1) We propose to forecast stock trends using LLMs which include ChatGPT and GPT-3 models under few-shot settings. 
This enables investors to forecast with minimal annotated financial data, making it more accessible to make timely precise decisions. 
(2) To facilitate few-shot stock trend prediction with LLMs, we further propose a two-step method `\textit{denoising-then-voting}' to reduce the adverse impact of noisy news and alleviate the input limitation of LLMs.
(3) We conduct extensive experiments on three datasets including S\&P 500 index, CSI-100 index, and HK stocks. Experimental results demonstrate the superior performance of our proposed method over few-shot LLM baselines. More importantly, when compared to state-of-the-art fully supervised methods, our approach exhibits competitive performances, verifying its effectiveness in stock trend prediction.

\section{Preliminary}
\subsection{Problem Definition}
Following the previous work~\cite{du2020stock,huang2022news}, we formulate stock trend prediction as a binary classification task. Given a target stock index $s$ and a specific date $k$, we aim to predict price movements $y_k \in \{Up, Down\}$ for stock $s$ on the $k$-\textit{th} day based on the $k-1$-\textit{th} day's news information $D_{k-1}$, which contains $|D_{k-1}|$ pieces of news. 
The predicted labeled $y_k=Up$ indicates the adjusted closing price of stock $s$ on the $k$-\textit{th} day will be greater than that on the previous day (i.e., $k-1$-\textit{th} day), while $y_k=Down$ represents the stock price will be less or equal than the previous day. 

\begin{figure*}[!t]
\centering
\includegraphics[width=\textwidth]{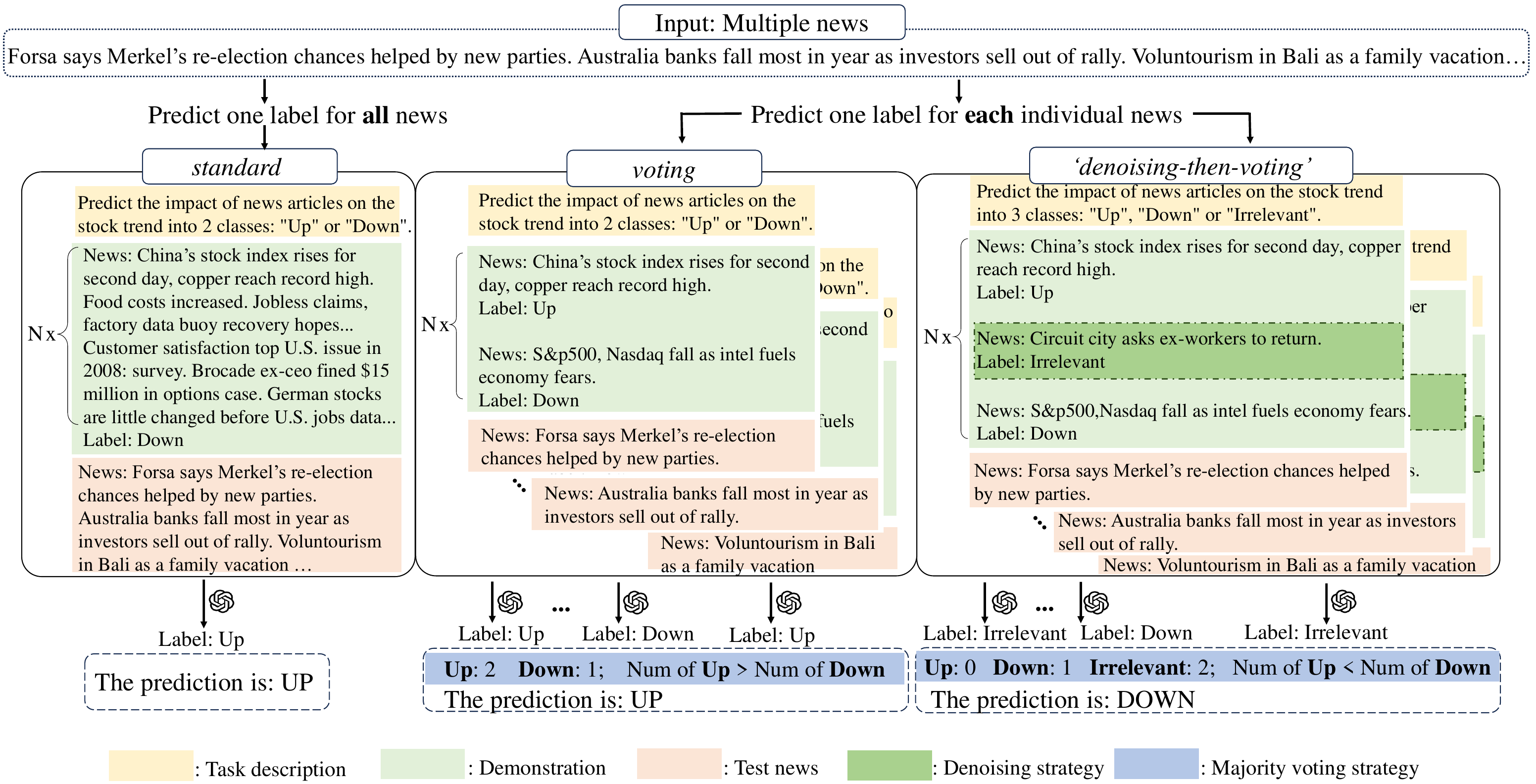}
\caption{An overview of three prompts used by LLMs in stock trend prediction. 
On the right, we display our `\textit{denoising-then-voting}' paradigm. The
heavy green part is the denoising strategy, where we introduce an ‘Irrelevant’ category and process each news individually. 
The blue part is the majority voting strategy where we reuse the predicted label of each individual news from the denoising stage and obtain the final predicted result through a majority voting mechanism.
The \textit{standard} and \textit{voting} prompts are respectively displayed on the left and in the middle for comparison.
} 
\label{fig:framework}
\end{figure*}

\subsection{In-context Learning}
Recent work~\cite{brown2020language,wei2022chain} has shown that pre-trained LLMs are strong in-context learners in few-shot scenarios across various NLP tasks even without additional fine-tuning or gradient updates. 
To conduct in-context learning, LLMs are supplied with a prompt comprising natural language instructions and a small set of exemplars pertaining to the target task.

\section{Approach}
\subsection{Overview}
Figure~\ref{fig:framework} depicts an overview of three different prompts used by LLMs. 
Our proposed`\textit{denoising-then-voting}' method is outlined on the right. 
It contains two components: a denoising strategy and a majority voting strategy. 
The denoising strategy (please see \S \ref{sec.denoising}) reduces irrelevant news and the majority voting strategy (please see \S \ref{sec.voting}) effectively aggregates relevant news for prediction.
By comparison, we also set up a \textit{standard} prompt on the left and a \textit{voting} prompt without denoising in the middle.
To predict stock trends, 
the standard way is to adopt a similar input manner as in supervised methods, which merge multiple news from the previous days and directly predict a final result for merged news. 
Therefore, in a \textit{standard} prompt, each exemplar consists of financial news paired with a corresponding label of stock trend.

\subsection{Denoising Strategy} \label{sec.denoising}
As previously mentioned, features from noise would hinder making accurate decisions~\cite{huang2022news,du2024financial}.
To filter out irrelevant news (i.e., reducing noise), we introduce an additional category named `Irrelevant' and predict on each news individually instead of on the merged news. 
Specifically, LLMs are required to classify each piece of news as `Irrelevant', `Up', or `Down', depending on whether it is unrelated to stock trends or contains relevant information that affects stock prices positively or negatively. 
By processing each news separately, 
 we can effectively identify and exclude irrelevant news, ensuring it does not interfere with our final stock trend prediction.

\subsection{Majority Voting Strategy} \label{sec.voting}

To utilize the remaining relevant news, 
one straight approach is to consolidate all these news into a single merged news and use LLMs to predict on the merged news. 
However, merging news may exceed the maximum token limits of LLMs, 
while cutting the combined news to fit the input token limits can lead to a loss of useful information. 
Furthermore, adopting this strategy necessitates using LLMs twice to obtain the final prediction – once during the denoising stage and once in the final prediction stage, which introduces additional costs and latency.
To bypass these problems, we opt to reuse the predicted results of individual news in the denoising stage and obtain a final prediction through an effective and efficient majority voting mechanism. 
As illustrated in the right part of Figure \ref{fig:framework},
the number of news classified as `Down' ($N_{down}$) outweighs those classified as `Up' ($N_{up}$) after excluding irrelevant news. 
Here we set a threshold $\lambda$, 
when $N_{down}/ (N_{up}+N_{down})> \lambda$, the final predicted stock trend is `Down', otherwise it is `Up'. 

\section{Experiments}

\subsection{Experimental Setups}
\subsubsection{Dataset.} 
We conduct experiments on three 
datasets: 
\textbf{\textit{U.S. S\&P 500}} involves 553,666 financial news texts (i.e., news titles and articles) released by~\citet{ding2015deep}.
The news texts sourced from \textit{Bloomberg and Reuters} are related to the U.S. stock market between 10/2006 to 11/2013.
Following~\citet{huang2022news}, we collect historical prices of the Standard \& Poor’s 500 (S\&P 500) index within the same period from \textit{Yahoo finance}\footnote{\url{https://finance.yahoo.com/}}, and derive the U.S. S\&P 500 dataset. 
\textbf{\textit{CSI-100 \& HK}} includes 90,361 financial news (i.e., news titles)\footnote{\url{https://pan.baidu.com/s/1mhCLJJi, 2017}} of 78 A-share stocks in CSI 100 and 13 Hong Kong stocks over the period from January 2015 to December 2015. 
Following~\citet{huang2018tensor}, we separately collect prices of the CSI-100 index, HK stocks from \textit{Wind}\footnote{\url{https://www.wind.com.cn/}}, and obtain the CSI-100 and HK dataset. Table~\ref{table:dataset stats} shows the statistics of these datasets.

\subsubsection{Evaluation Metrics.} 
In line with the previous work~\cite{huang2018tensor,huang2022news}, 
we employ four established classification metrics: Precision (P), Recall (R), F1-score (F1), and Accuracy (Acc) to evaluate the performance of the stock trend prediction task.

\subsubsection{Baselines.} 
We compare our proposed `\textit{denoising-then-voting}' method with the few-shot baselines and state-of-the-art supervised baselines.
\textbf{\textit{Few-Shot Baselines.}}
We take the \textit{standard} prompt that predicts stock trends for the merged news, and the \textit{voting} prompt that predicts on each individual news instead of the merged news to vote. The `Irrelevant' label (for denoising purpose) is not introduced in these baselines. 
Both few-shot baselines and our proposed method use the same LLMs for fair comparisons. 
\textbf{\textit{Supervised Baselines.}}
We include several state-of-the-art supervised baselines, which utilize the merged news as input to forecast stock trends. 
In S\&P 500 evaluations, we compare models with different news representations. 
\textbf{CNN+KB}~\cite{ding-etal-2016-knowledge} uses event embeddings enhanced with knowledge bases and CNNs as the event sequence model. Later, \textbf{LSTM-RGCN}~\cite{li2021modeling} takes LSTM to encode news texts and Relational Graph Convolutional Network (RGCN) to learn node representations for each stock. 
The Noise Equity State representation of news (\textbf{NES}) in~\citet{huang2022news} integrates representations of equity state and noise effects into their model.
For CSI-100 \& HK, we compare our proposed method with: 
\textbf{TeSIA}~\cite{li2015tensor}, the tensor-based information framework in predicting stock movements, 
\textbf{GDR+TeSIA}~\cite{li2016tensor} and \textbf{SMC+TeSIA}~\cite{huang2018tensor}, which reduces input dimensionality for TeSIA 
using Global Dimensionality-Reduction (GDR) and Sub-mode Coordinate (SMC), 
and \textbf{SMC+LSTM}~\cite{huang2018tensor}, a variant of SMC+TeSIA that utilizes LSTM to predict stock trends.

\begin{table}[]
\footnotesize
\centering
\begin{tabular}{
m{0.117\textwidth}<{\raggedright}
m{0.087\textwidth}<{\raggedright}
m{0.087\textwidth}<{\raggedright}
m{0.087\textwidth}<{\raggedright}
}
\hline \specialrule{0em}{1pt}{1pt} 
\textbf{U.S. S\&P 500}   & Training                                                              & Validation                                                        & Test                                                               \\ \specialrule{0em}{1pt}{1pt} 
 \hline \specialrule{0em}{1pt}{1pt}
\#Tokens/news            & 219.93                                                             & 495.37                                                             & 455.58                                                             \\
\#Tokens/day          & 2093.35                                                            & 4550.38                                                            & 4179.95                                                            \\
\#Days              & 1425                                                               & 169                                                                & 191                                                                \\
Time span              & \begin{tabular}[c]{@{}l@{}}10/20/2006\\ --06/18/2012\end{tabular} & \begin{tabular}[c]{@{}l@{}}06/19/2012 \\ --02/21/2013\end{tabular} & \begin{tabular}[c]{@{}l@{}}02/22/2013 \\--11/21/2013\end{tabular} \\ 
Label distribution & Down / Up (647 / 778) & Down / Up (80 / 89) & Down / Up (88 / 103)\\ \specialrule{0em}{1pt}{1pt}\hline \specialrule{0em}{1pt}{1pt}
\textbf{CSI-100\&HK} & Training                                                              & Validation                                                        & Test                                                               \\\specialrule{0em}{1pt}{1pt} \hline \specialrule{0em}{1pt}{1pt}
\#Tokens/news            & 273.64                                                             & 237.40                                                             & 247.79                                                             \\
\#Tokens/day          & 6011.22                                                            & 4921.10                                                            & 5231.24                                                            \\
\#Days              & 164                                                                & 20                                                                 & 63                                                                 \\
Time span              & \begin{tabular}[c]{@{}l@{}}01/01/2015\\ --31/08/2015\end{tabular}  & \begin{tabular}[c]{@{}l@{}}01/09/2015\\--30/09/2015\end{tabular}  & \begin{tabular}[c]{@{}l@{}}01/10/2015\\--31/12/2015\end{tabular}  \\ 
Label distribution & Down / Up (80 / 84) & Down / Up (12 / 8) & Down / Up (37 / 26)\\ \specialrule{0em}{1pt}{1pt} \hline
\end{tabular}
\caption{Statistics of the datasets used in this paper. \#Tokens/news and \#Tokens/day denote the average number of tokens within one news title and one trading day. \#Days represents the number of trading days in training, validation, and test sets.}
\label{table:dataset stats}
\end{table}

\subsubsection{Implementation Details.} 
For both the few-shot baselines and our proposed `\textit{denoising-then-voting}' method, we utilize ChatGPT (i.e., gpt-3.5-turbo-0301\footnote{\url{https://platform.openai.com/docs/models/gpt-3-5}}) as the backbone in-context model and set the temperature to 0. 
Following previous works~\cite{huang2018tensor,huang2022news}, we leverage the title of news (i.e., news titles) to predict stock trends. Please refer to \S \ref{sec.input} for the effect of different news inputs (e.g., news titles or articles). 
To set up our `\textit{denoising-then-voting}' method, we adopt a 9-shot (3 examples per class) prompt for U.S. S\&P 500 dataset, and a 6-shot (2 examples per class) prompt for CSI and HK datasets. 
For the \textit{standard} and \textit{voting} prompts, we also use the same settings. 
(please refer to Appendix \ref{sec:appendix_prompt_construction} for more details.)
We choose threshold $\lambda$ that performs best in the validation set and evaluate it on the test set. Here, we take $\lambda=\frac{1}{2}$ on the three datasets.

\begin{table}[!t]
\footnotesize
\centering
\begin{tabular}{
m{0.2\textwidth}<{\raggedright}
m{0.035\textwidth}<{\centering}
m{0.035\textwidth}<{\centering}
m{0.035\textwidth}<{\centering}
m{0.035\textwidth}<{\centering}
}

\hline
\specialrule{0em}{1pt}{1pt}
\multirow{2}{*}{\textbf{Models}}    & \multicolumn{4}{c}{\textbf{S\&P 500}}                                         \\ \specialrule{0em}{0.5pt}{0.5pt}\cline{2-5} \specialrule{0em}{0.5pt}{0.5pt}
                                   & \textbf{Acc} & \textbf{P} & \textbf{R} & \textbf{F1} \\\specialrule{0em}{1pt}{1pt} \hline \specialrule{0em}{1pt}{1pt}
\multicolumn{5}{l}{\textbf{Fully Supervised Methods}}                                                     \\
CNN+KB                           & 66.93           & 68.63            & 57.35         & 62.49     \\
LSTM-RCCN                       & 63.87           & 69.04            & 59.34         & 63.82     \\
NES                                 & \underline{67.34}           & \underline{69.74}            & \underline{61.82}         & \underline{65.54}     \\  \specialrule{0em}{1pt}{1pt} \hline \specialrule{0em}{1pt}{1pt} 
\multicolumn{5}{l}{\textbf{Few-Shot (\textit{standard})}}                                                             \\
6-shot ChatGPT (20)          & 59.75           & 71.38            & 62.01         & 65.73\\ \specialrule{0em}{1pt}{1pt} \hline \specialrule{0em}{1pt}{1pt}
\multicolumn{5}{l}{\textbf{Few-Shot (\textit{\textit{voting}})}}                                                             \\
6-shot ChatGPT (60)          &  62.36           & 86.79           & 61.90        & 73.21  \\
8-shot ChatGPT (60)          &  62.35           & 83.77               & 62.06        & 72.94 \\
10-shot ChatGPT (60)          & 60.74            & 82.89             & 60.60        & 71.96 \\
\specialrule{0em}{1pt}{1pt} \hline \specialrule{0em}{1pt}{1pt}
\multicolumn{5}{l}{\textbf{Few-Shot (`\textit{denoising-then-voting}')}}                                                           \\
9-shot ChatGPT (20)          & 61.09           & 81.54            & 62.84         & 70.12\\ 
9-shot ChatGPT (60) & \textbf{66.59}           & \textbf{94.17}            & \textbf{66.90}         & \textbf{75.27} \\ \specialrule{0em}{1pt}{1pt}  \hline   
\end{tabular}
\caption{The overall performance (\%) on U.S. S\&P 500 dataset. 
We \textbf{bold} the best results for few-shot methods and \underline{underline} the highest results for supervised methods. The numbers in brackets denote the number of news titles utilized for forecasting stock trends. 
}
\label{table:overall_sp500}
\end{table}

\begin{table}[!t]
\footnotesize
\centering
\begin{tabular}
{
m{0.25\textwidth}<{\raggedright}
m{0.09\textwidth}<{\centering}
m{0.06\textwidth}<{\centering}
}
\specialrule{0em}{1pt}{1pt} 
\hline
\specialrule{0em}{1pt}{1pt} 
\textbf{Model}                       & \textbf{CSI-100}          & \textbf{HK}               \\ \specialrule{0em}{1pt}{1pt}  \hline \specialrule{0em}{1pt}{1pt} 
\multicolumn{3}{l}{\textbf{Fully Supervised Methods}}                                                                                                                                                                 \\
TeSIA                                & 60.63          & 60.38          \\
GDR+TeSIA                            & 61.46          & 60.92          \\
SMC+TeSIA                             & 62.29          & 62.05          \\
SMC+LSTM                              & \underline{62.97}    & \underline{63.31}    \\ \specialrule{0em}{1pt}{1pt} \hline \specialrule{0em}{1pt}{1pt} 
\multicolumn{3}{l}{\textbf{Few-Shot (\textit{standard})}}                                                                                                                                                                 \\
4-shot ChatGPT (10)                         & 58.17          & 57.01          \\ \specialrule{0em}{1pt}{1pt} \hline\specialrule{0em}{1pt}{1pt}
\multicolumn{3}{l}{\textbf{Few-Shot (\textit{voting})}}                                                                                                                                                                 \\
4-shot ChatGPT (20)                         &  58.42           &  57.92          \\
6-shot ChatGPT (20)                         &  57.92         &  55.41          \\ \specialrule{0em}{1pt}{1pt} \hline\specialrule{0em}{1pt}{1pt}
\multicolumn{3}{l}{\textbf{Few-Shot (`\textit{denoising-then-voting}')}}                                                                                                                                                                 \\
6-shot ChatGPT (10)                         & 57.25          & 56.75          \\
6-shot ChatGPT (20)                         & \textbf{62.17} & \textbf{61.17} \\ \specialrule{0em}{1pt}{1pt} \hline \specialrule{0em}{1pt}{1pt} 
\end{tabular}
\caption{The overall performance in accuracy (\%) on CSI-100 and HK datasets. We \textbf{bold} the best results for few-shot methods and \underline{underline} the highest results for supervised methods. 
}
\label{table:overall_CSI-HK}
\end{table}

\subsection{Overall Performance} 
We show the overall results of our proposed method for S\&P 500 index in Table~\ref{table:overall_sp500}, and CSI-100 index and HK stocks in Table~\ref{table:overall_CSI-HK}. 
In Table~\ref{table:overall_sp500}, our method yields a prediction accuracy of 66.59\% for U.S. S\&P 500 forecasting, with a 6.84\% enhancement over the ChatGPT with the \textit{standard} prompt (i.e., 6-shot ChatGPT). 
Similarly, Table~\ref{table:overall_CSI-HK} shows that our approach achieves prediction accuracy of 62.17\% and 61.17\% for the CSI-100 and HK stock respectively, demonstrating 4\% improvements compared to the \textit{standard} prompt (i.e., 4-shot ChatGPT). 
In addition, our method obtains better performance compared with the \textit{voting} prompt, with 3\%-4\% elevations in accuracy across three datasets. All improvements are statistically significant compared with the \textit{standard} and \textit{voting} few-shot methods, with \textit{$p$}-values less than 0.05 under the paired t-test.

The major improvements of our proposed `\textit{denoising-then-voting}' method come from two aspects. 
The denoising strategy helps remove noise, enhancing performance by focusing on relevant information.
Besides, the majority voting strategy effectively aggregates financial news. 
In Table~\ref{table:overall_sp500} and~\ref{table:overall_CSI-HK}, 
much more news can be utilized in prediction than the \textit{standard} prompt.
So, the majority voting strategy mitigates the loss of useful information about stock trends and further elevates accuracy.

Overall, our proposed `\textit{denoising-then-voting}' method outperforms both \textit{standard} and \textit{voting} few-shot approaches on different datasets. Furthermore, it yields results comparable to the state-of-the-art supervised methods across three datasets, highlighting the effectiveness of our `\textit{denoising-then-voting}' method for stock trend prediction tasks under few-shot settings.

\subsection{Ablation Study}
To assess the effects of different components within our proposed method, in this section, we conduct an ablation study and obtain results in Table~\ref{table:ablations}.
Compared to the \textit{standard} prompt, the majority voting strategy yields a significant improvement in accuracy (2.61\%) on the S\&P 500 dataset, 
by mitigating the loss of news information.
However, the benefits of majority voting are more modest on the CSI-100 and HK datasets 
because these two datasets contain more irrelevant news. 
Therefore, we propose a denoising strategy by introducing an `Irrelevant' label to distinguish noise from informative news,  deriving the proposed `\textit{denoising-then-voting}' method. 
Table~\ref{table:ablations} demonstrates that the `\textit{denoising-then-voting}' method outperforms the \textit{voting} prompt by a large margin, achieving an accuracy increase of 3\%-4\% on all three datasets. 
These improvements highlight the benefits of the denoising strategy, as it effectively identifies irrelevant news, and mitigates its adverse impacts on the final prediction. 
To summarize, both the denoising and the majority voting are indispensable to our proposed method for stock trend prediction.

\begin{table}[]
\centering
\footnotesize
\begin{tabular}
{
m{0.155\textwidth}<{\raggedright}
m{0.08\textwidth}<{\centering}
m{0.075\textwidth}<{\centering}
m{0.065\textwidth}<{\centering}
}
\hline
\specialrule{0em}{0.7pt}{0.7pt}
\multicolumn{1}{l}{\textbf{ChatGPT}}      & \textbf{S\&P 500} & \textbf{CSI-100} & \textbf{HK}    \\ \specialrule{0em}{0.5pt}{0.5pt}  \hline \specialrule{0em}{0.7pt}{0.7pt} 
\textit{standard}                         & 59.75    & 58.17   & 57.01 \\ 
\textit{voting}                & 62.36    & 58.42   & 57.92 \\ \specialrule{0em}{0.5pt}{0.5pt}  \hline \specialrule{0em}{0.7pt}{0.7pt} 
\textit{`D+V'} & \textbf{66.59}    & \textbf{62.17}   & \textbf{61.17} \\\specialrule{0em}{0.5pt}{0.5pt}  \hline \specialrule{0em}{0.5pt}{0.5pt} 
\end{tabular}
\caption{Ablation accuracy (\%) on different datasets. For the \textit{standard} prompt, we report the best performance it gains before reaching the context window limits (4,097 tokens) of ChatGPT. 
`D+V' represents our `\textit{denoising-then-voting}' method.
}
\label{table:ablations}
\end{table}

\subsection{Further Analysis}
We further evaluate our `\textit{denoising-then-voting}' method in various experimental settings such as using different LLMs, inputting different number of news titles, etc. By default, we use ChatGPT and employees 60 news titles under 9-shot setting (3 examples per class) in S\&P 500 prediction, 40 titles in 6-shot setting (2 examples per class) for CSI and HK predictions.
\subsubsection{Performance on Different LLMs.}
\begin{figure}[!t]
\centering
\includegraphics[width=0.5\textwidth]{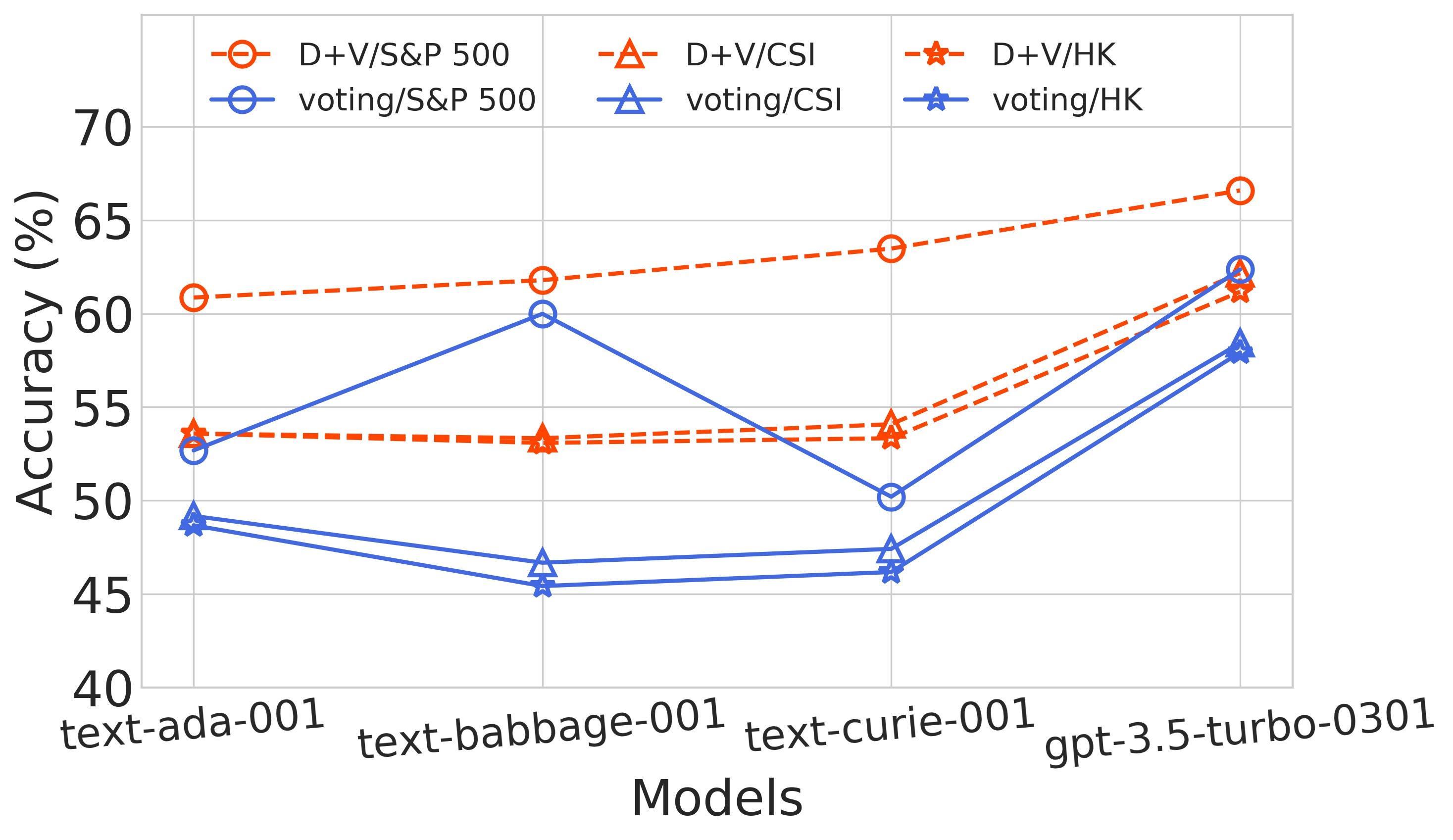}
\caption{Comparison between the \textit{voting} and `\textit{denoising-then-voting}' (D+V) methods on S\&P 500, CSI-100 and HK across different LLMs.} 
\label{fig:diffrent llm}
\end{figure}

Besides ChatGPT, we experiment with three different sizes of GPT-3 models~\cite{brown2020language}: GPT-3 Ada (350 million parameters), GPT-3 Babbage (3 billion parameters), and GPT-3 curie (13 billion parameters). 
As shown in Figure~\ref{fig:diffrent llm}, the `\textit{denoising-then-voting}' method consistently outperforms the \textit{voting} prompt across various LLMs on all three datasets, demonstrating the robustness and superior performance of our proposed approach. 

\subsubsection{Effects of the Number of Exemplars.} 
To assess the effect of the number of exemplars in our proposed `\textit{denoising-then-voting}' method, we experiment on prompts with various numbers of exemplars (i.e., $k$-shot in-context learning). As shown in Table~\ref{table:k-shot selection}, our proposed `\textit{denoising-then-voting}' method performs best under 9-shot (3 examples per class) setting in S\&P 500 index prediction and 6-shot (2 examples per class) setting for CSI-100 and HK stock forecast. 
With few exemplars (e.g., 0-shot), LLMs are not provided with sufficient examples, which may limit their capability of capturing information.  
When given too many exemplars (12-shot for S\&P 500, 9-shot for CSI and HK), however, the performance of our proposed method may become saturated or even decrease. 
Therefore, in this work, we adopt a 9-shot prompt for U.S. S\&P 500 dataset and a 6-shot prompt for both CSI-100 and HK datasets.

\begin{table}[!t]
\footnotesize
\centering
\begin{tabular}{lccccc}
\hline
\specialrule{0em}{1pt}{1pt}
\multicolumn{1}{l}{\textbf{Dataset}} & \textbf{0-shot} & \textbf{3-shot} & \textbf{6-shot} & \textbf{9-shot} & \textbf{12-shot} \\ \specialrule{0em}{1pt}{1pt}\hline\specialrule{0em}{1pt}{1pt}
S\&P 500                             & 63.67         & 64.70         & 64.73        & \textbf{66.59}         & 63.96          \\
CSI-100                              & 58.67         & 61.17         & \textbf{62.17}         & 58.25         & -                \\
HK                                   & 58.17         & 59.01         & \textbf{61.17}         & 57.25         & -                \\ \hline
\end{tabular}
\caption{Accuracy (\%) for various numbers of exemplars in in-context learning.}
\label{table:k-shot selection}
\end{table}

\begin{table*}[t]
\centering
\footnotesize
\begin{tabular}{
m{0.3cm}<{\raggedright} 
m{4.8cm}<{\raggedright} 
m{4.cm}<{\raggedright} 
m{4.5cm}<{\raggedright}}
\hline
\specialrule{0em}{1pt}{1pt}
& \multicolumn{1}{c}{\textbf{`\textit{denoising-then-voting}'}}                                                                                                                                                                                                                                                                                                                                                                                               & \multicolumn{1}{c}{\textbf{\textit{voting}}}                                                                                                                                                                                                                                                                                                                                                                                               & \multicolumn{1}{c}{\textbf{\textit{standard}}}                                                                                                                                                                                                                                                                                                           \\ \specialrule{0em}{1pt}{1pt} \hline \specialrule{0em}{1pt}{1pt}
\multirow{9}{*}{\#1} & \multicolumn{2}{l}{\textbf{News 1}: Forsa says merkel's re-election chances helped by new parties. } & \multirow{8}{*}{\begin{tabular}[c]{@{}l@{}}\textbf{News}: Forsa says merkel's re-\\election chances helped by new\\parties...Australia banks fall most\\in year as investors sell out of rally.\\...Voluntourism in Bali as a family\\vacation...China starts website to\\refute rumors as scrutiny grows...\end{tabular}} \\
& \textbf{Pred}: \underline{Irrelevant}                                     & \textbf{Pred}: \underline{Up}                            &                                                                                                                                                                                                                                                                                                                      \\
& \multicolumn{2}{l}{\textbf{News 2}: Australia banks fall most in year as investors sell out of rally.} &                                                                                                                                                                                                                                                                                                                      \\
& \textbf{Pred}: \underline{Down}                                           & \textbf{Pred}: \underline{Down}                            &                                                                                                                                                                                                                                                                                                                      \\
& \multicolumn{2}{l}{\textbf{News 3}: Voluntourism in Bail as a family vacation.}                        &                                                                                                                                                                                                                                                                                                                      \\
& \textbf{Pred}: \underline{Irrelevant}                                     & \textbf{Pred}: \underline{Up}                              &                                                                                                                                                                                                                                                                                                                      \\
& \multicolumn{2}{l}{\textbf{News 4}: China starts website to refute rumors as scrutiny grows.}                   &                                                                                                                                                                                                                                                                                                                      \\
& \textbf{Pred}: \underline{Irrelevant}                                     & \textbf{Pred}: \underline{Up}                              &                                                                                                                                                                                                                                                                                                                      \\
                                                      & \textbf{\textcolor{blue}{Final: Down \quad \quad Actual: Down}}  &  \textbf{\textcolor{red}{Final: Up \quad \  Actual: Down}}  & \textbf{\textcolor{red}{Final:  \underline{Up} \quad \quad \quad Actual: Down}} \\\specialrule{0em}{1pt}{1pt} \hline \specialrule{0em}{1pt}{1pt}

\multirow{9}{*}{\#2} & \multicolumn{2}{l}{\textbf{News 1}: Fed bridges gap to earnings pickup in modest u.s. growth. } & \multirow{8}{*}{\begin{tabular}[c]{@{}l@{}}\textbf{News}: Fed bridges gap to earnings\\ pickup in modest u.s. growth...\\Treasury yields at six-week high\\as gross says bull over... \textcolor{gray}{Lone}\\\textcolor{gray}{star said to finish raising \$5 billion}\\\textcolor{gray}{property fund... Japan current-}\\\textcolor{gray}{account surplus climbs as }\\\textcolor{gray}{abenomics sinks yen...} \end{tabular}} \\ 
& \textbf{Pred}: \underline{Irrelevant}                                     & \textbf{Pred}: \underline{Up}                            &                                                                                                                                                                                                                                                                                                                      \\
& \multicolumn{2}{l}{\textbf{News 2}: Treasury yields at six-week high as gross says bull over.} &                                                                                                                                                                                                                                                                                                                      \\
& \textbf{Pred}: \underline{Down}                                           & \textbf{Pred}: \underline{Down}                            &                                                                                                                                                                                                                                                                                                                      \\
& \multicolumn{2}{l}{\textbf{News 3}: Lone star said to finish raising \$5 billion property fund.}                        &                                                                                                                                                                                                                                                                                                                      \\
& \textbf{Pred}: \underline{Up}                                     & \textbf{Pred}: \underline{Up}                              &                                                                                                                                                                                                                                                                                                                      \\
& \multicolumn{2}{l}{\textbf{News 4}: Japan’s current-account surplus climbs as abenomics sinks yen. }                   &                                                                                                                                                                                                                                                                                                                      \\
& \textbf{Pred}: \underline{Up}                                     & \textbf{Pred}: \underline{Up}                              &                                                                                                                                                                                                                                                                                                                      \\
                                                      & \textbf{\textcolor{blue}{Final: Up \quad \quad \quad Actual: Up}}  &  \textbf{\textcolor{blue}{Final: Up \quad \quad Actual: Up}}  & \textbf{\textcolor{red}{Final:  \underline{Down} \quad \quad \quad Actual: Up}} \\\specialrule{0em}{1pt}{1pt} \hline \specialrule{0em}{1pt}{1pt}
\textbf{Acc} & \multicolumn{1}{c}{\textbf{66.59\%}}                                                                                                                                                                                                                                                                                                                                                                                                                                     & \multicolumn{1}{c}{\textbf{62.36\%}}                                                                                                                                                                                                                                                                                                                                                                                                                      & \multicolumn{1}{c}{\textbf{59.75\%}}                                                                                                                                                                                                                                                                                                                            \\\specialrule{0em}{1pt}{1pt} \hline
\end{tabular}
\caption{Stock trends predicted by the `\textit{denoising-then-voting}', \textit{voting} and \textit{standard} methods on U.S. S\&P 500 dataset. We \underline{underline} the output generated by ChatGPT. \textcolor{blue}{Blue text} represents the final prediction consistent with the ground truth. \textcolor{red}{Red text} denotes the final prediction that contradicts the real stock movement. \textcolor{gray}{Gray text} is the text beyond the input limit of LLMs. For the sake of simplicity, here we only present 4 news texts as the long text input and display outputs obtained from in-context learning. }
\label{tabel:case_study}
\end{table*}

\begin{figure}[!t]
\centering
\includegraphics[width=0.5\textwidth]{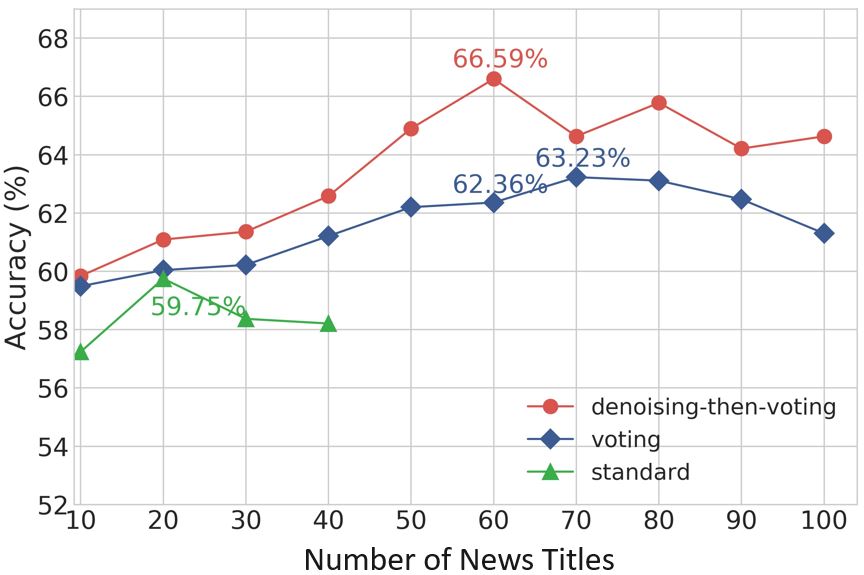}
\caption{The prediction accuracy of S\&P 500 index when aggregating various numbers of news titles with three few-shot methods.}
\label{fig:random_news_number_sp}    
\end{figure}

\subsubsection{Effects of the Number of News Titles.}  
We input different numbers of news titles in predicting S\&P 500 movement for evaluation.
As depicted in Figure \ref{fig:random_news_number_sp},
when inputting various numbers of news titles, our method exhibits a consistent enhancement compared to the \textit{standard} and \textit{voting} prompts. 
The \textit{standard} prompt can only accommodate up to 40 news titles at one time for the input token limits of ChatGPT.
In contrast, the \textit{voting} prompt can accept much more news titles. Accordingly, the \textit{voting} prompt outperforms the \textit{standard} prompt by a large margin (63.23\% vs. 59.75\%). 
More importantly, our `\textit{denoising-then-voting}' method further significantly improves the accuracy (66.59\% vs. 63.23\%) over the \textit{voting} prompt, 
showing the effectiveness of denoising strategy.

\subsubsection{Effects of Inputting Titles or Articles.} \label{sec.input}

\begin{table}[!t]
\footnotesize
\centering
\begin{tabular}{
m{0.28\textwidth}<{\raggedright}
m{0.14\textwidth}<{\centering}
}
\specialrule{0em}{1pt}{1pt}  \hline \specialrule{0em}{1pt}{1pt} 
\textbf{Varied inputs}               & \textbf{Accuracy}         \\ \specialrule{0em}{1pt}{1pt}  \hline \specialrule{0em}{1pt}{1pt} 
\multicolumn{2}{l}{\textbf{\textit{standard}}}           \\
Title            & 59.75\%          \\
Article-last-100    & 60.80\%          \\
Article-middle-100 & 61.85\%          \\
Article-first-100   & \underline{63.52\%}    \\
Article-summary-100   & 61.30\%    \\ \specialrule{0em}{1pt}{1pt}  \hline \specialrule{0em}{1pt}{1pt}  
\multicolumn{2}{l}{\textbf{`\textit{denoising-then-voting}'}}  \\
Title             & 61.09\%          \\
Article-first-10             & 59.24\%\\
Article-first-100   & \textbf{64.07\%} \\ \specialrule{0em}{1pt}{1pt}  \hline \specialrule{0em}{1pt}{1pt} 
\end{tabular}
\caption{Performance comparison when inputting news titles or different parts of articles. Considering the inference cost of OpenAI's ChatGPT API, we use 20 titles or articles for S\&P 500 index prediction.}  
\label{table: passage_or_title}
\end{table}

Typically, a complete news article is quite lengthy. 
As news titles highly summarize the news articles, in this paper, we mostly use news titles as input to predict trends. 
In this section, we explore the effects of using news titles or articles as input for S\&P 500 forecast. 
Given that inputting a complete news article into LLMs would be costly, 
we propose four methods to condense a news article into a 100-token summary for prediction.
Article-first-100, Article-middle-100, and Article-last-100 respectively represent taking the first, the middle, and the last 100 tokens of the news article as the summary. Article-summary-100 refers to compressing the given news article into a summary of 100 tokens using ChatGPT\footnote{We apply the prompt provided by~\citet{zhang2023benchmarking} to ChatGPT for news summarization.}.
The average length of each news title is around 10 tokens. 
We also use Article-first-10, which takes the first 10 tokens of the news article as the summary, to compare with using 10-token news titles as input.

As shown in the first block of Table~\ref{table: passage_or_title}, for the \textit{standard} prompt, inputting with the first 100 tokens (Article-first-100) performs best than those inputting with the middle 100 tokens, the last 100 tokens and the summary generated by ChatGPT. That is because news typically has lead bias~\cite{zhang2020pegasus,zhu2021leveraging}, where the early parts of an article often contain the most salient information.  
In contrast, using the article's last 100 tokens as input performs the worst, indicating that the last 100 tokens contain less useful information. 

For `\textit{denoising-then-voting}', we find that employing the first 10 tokens (Article-first-10) as input performs worse than using news titles for stock trend prediction, which clearly demonstrates that news titles are highly condensed and high-quality summaries.
Conversely, using the first 100 tokens (Article-first-100) as input brings a significant improvement to the proposed `\textit{denoising-then-voting}' over using the news title as input. 
Since Article-first-100 takes much longer inputs than news titles, it includes more useful information for stock trend prediction. 
It is worth noting that inputting a lengthier summary also incurs higher API usage costs, which is often prohibitive in practice. Therefore, using news titles as input strikes a good balance between performance and cost efficiency.

\subsection{Case Study}
We present two cases of stock trend prediction in Table~\ref{tabel:case_study}. 
In case \#1, both the \textit{standard} and \textit{voting} prompts fail to identify noise (News 1, 3, 4), thus resulting in an incorrect final prediction. 
In contrast, our proposed method can exclude irrelevant news (News 1, 3, 4) and mitigate the adverse impacts of noise, leading to a correct final prediction based on the valuable information (News 2). Case \#1 highlights the denoising strategy. 
For case \#2, both \textit{voting} and `\textit{denoising-then-voting}' methods correctly predict the stock trends, while the \textit{standard} prompt does not. 
To fit the input limits of LLMs, meaningful information (News 3 and 4) is excluded when truncating the merged news, leading to a wrong prediction with the \textit{standard} prompt.  
Yet, the majority voting strategy allows us to effectively aggregate more news for prediction analysis. Therefore, it can include more information relevant to stock trends (News 3 and 4), elucidating the effectiveness of the majority voting strategy.

\section{Related Work}
\paragraph{News-based Stock Prediction} investigates the relationship between news data and stock market fluctuations.
To capture meaningful news information, 
various efficient news representations are explored, 
e.g., tensor-based method~\cite{huang2018tensor}, 
stock embeddings~\cite{du2020stock}, etc. 
In addition, deep learning methods~\cite{hu2021stock,zhao2023deep} have been developed to extract relevant patterns 
within financial data. 
Trained with extensive annotated data, these works accumulate multiple news together to predict their influence on the future market. 
However, the labeled data is not always available.
Instead, LLMs show remarkable few-shot capability while merging multiple news as input will cause input overflow of LLMs and introduce significant noise, harming prediction results.
To handle these issues, this paper proposes a `\textit{denoising-then-voting}' method and predicts trends with LLMs under a few-shot setting.

\paragraph{Large Language Models (LLMs)} have demonstrated remarkable capability in various text generation tasks, such as machine translation~\cite{jiao2023chatgpt}, question-answering~\cite{lin-etal-2022-truthfulqa} and factual error correction \cite{he-etal-2023-pivotfec, he2024improving}. 
In financial domains, 
the applications of LLMs still remain relatively untapped.
\citet{wu2023bloomberggpt} proposes BloombergGPT, a domain-specific LLM for finance. 
\citet{lopez2023can} 
scores news headlines using ChatGPT and conducts different trading strategies based on the scores to predict stock returns. Different from prior studies, this work investigates various few-shot methods. We contribute to alleviating the noise and input limitations with LLMs when processing thousands of pieces of news in stock trend prediction. 

\section{Conclusion}
In this work, we propose to use large language models (LLMs) such as ChatGPT and GPT-3 in stock prediction with financial news. It addresses the issue of sparse data annotation in supervised learning. More importantly, the proposed two-step method `\textit{denoising-then-voting}' effectively deals with noisy news and alleviates the constraints of LLMs in handling large input news streams. The results show significant improvements compared with few-shot baselines and are highly comparable to the state-of-the-art fully supervised methods. 
This study demonstrates the potential feasibility of adopting LLMs in financial forecasting with our proposed method, 
paving the way for truly enabling LLM applications 
in real-world trading.

\bibliography{acl_latex}

\appendix

\section{Prompt Construction}
\label{sec:appendix_prompt_construction}

\subsection{Full Few-Shot Prompts for S\&P 500 Prediction} \label{append:sp_prompts}

\begin{sidewaysfigure*}[!t]
\centering
\includegraphics[width=1\textwidth]{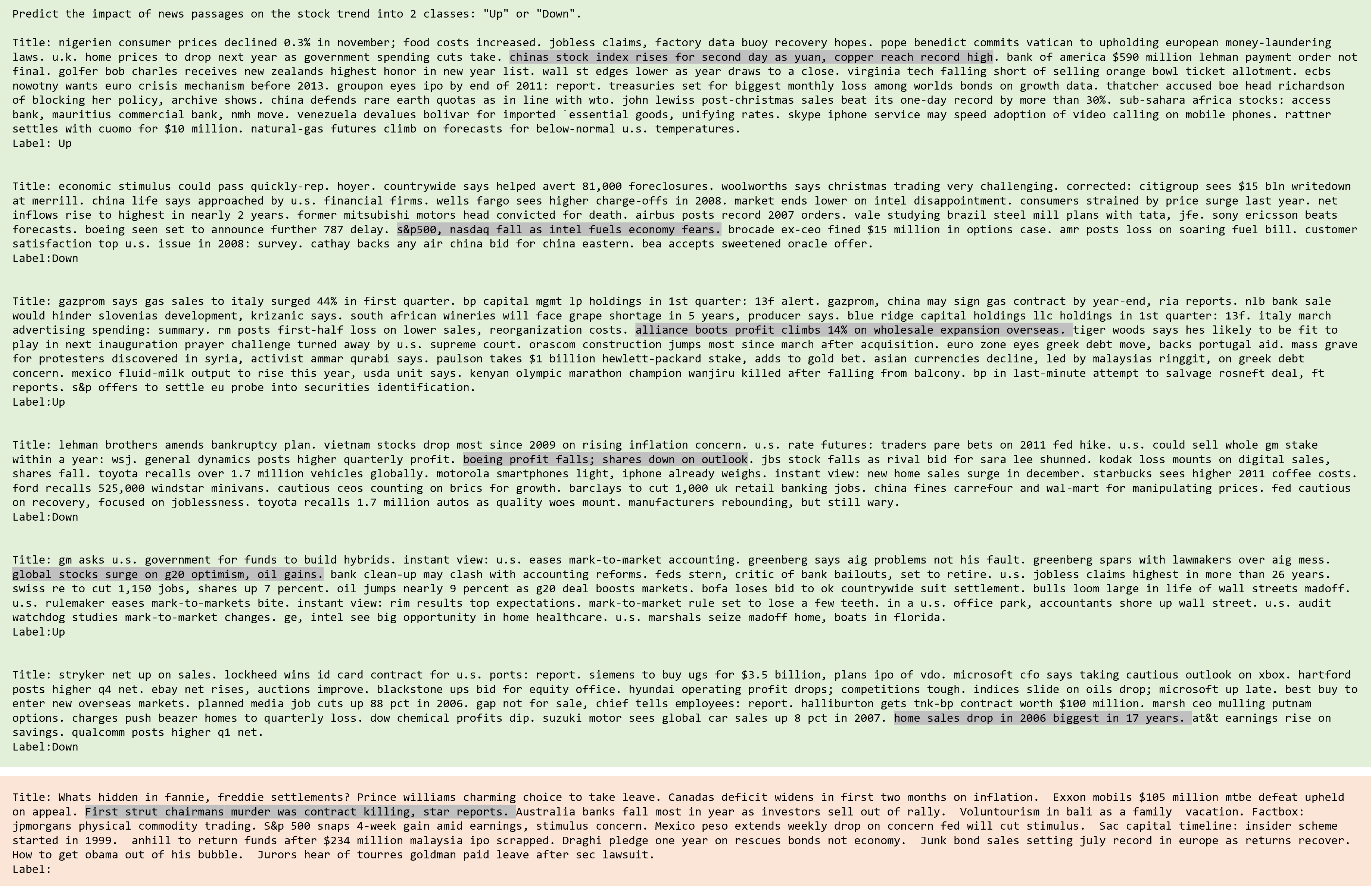}
\caption{The \textit{standard} prompt for S\&P 500 index prediction (20 titles).}
\label{fig:append_sp_standard}
\end{sidewaysfigure*}

\begin{figure*}[!t]
\centering
\includegraphics[width=0.55\textwidth]{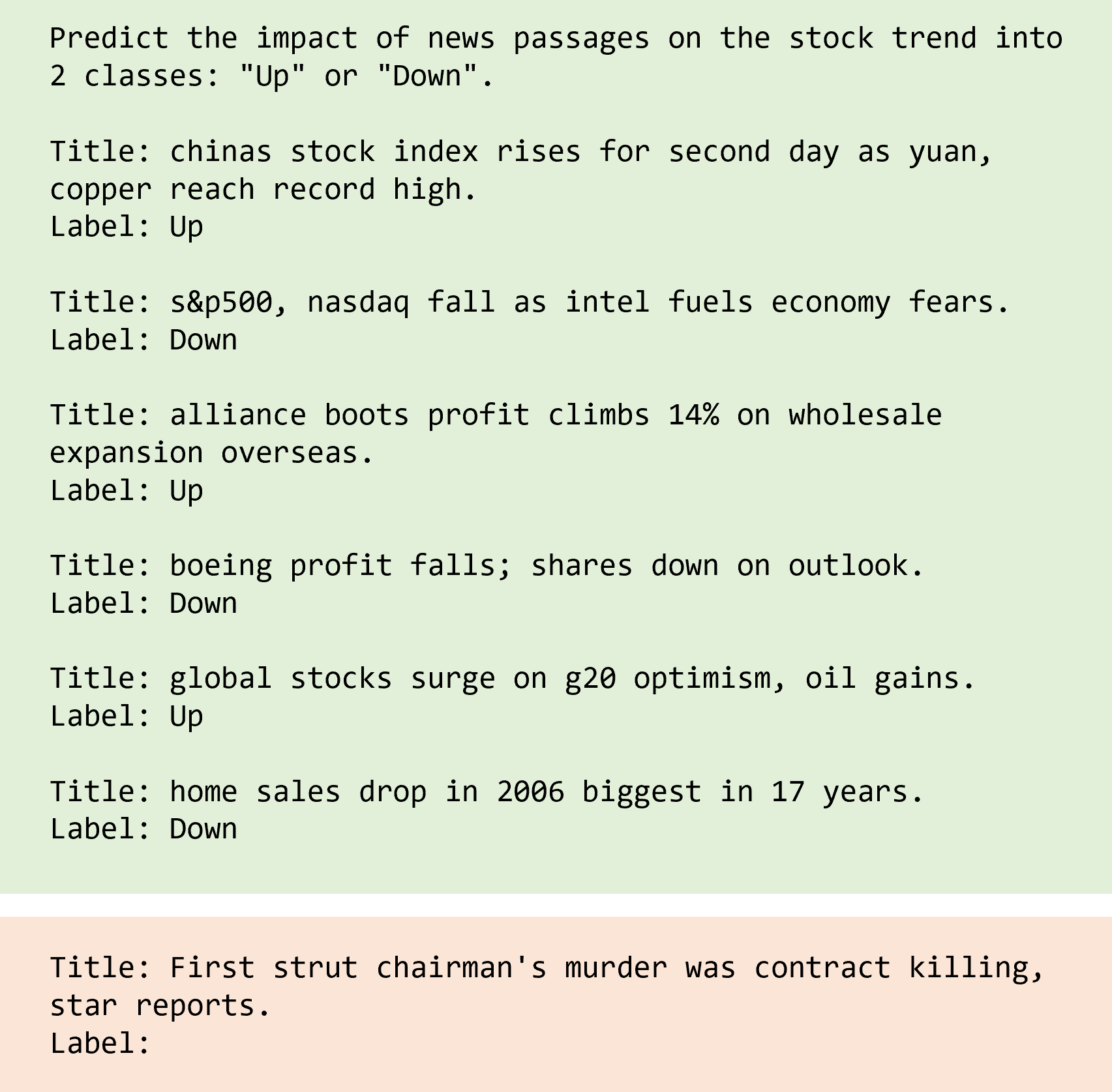}
\caption{The \textit{voting} prompt w/o `Irrelevant' category in S\&P 500 index prediction.}
\label{fig:append_sp_2vote}
\end{figure*}

\begin{figure*}[!t]
\centering
\includegraphics[width=0.65\textwidth]{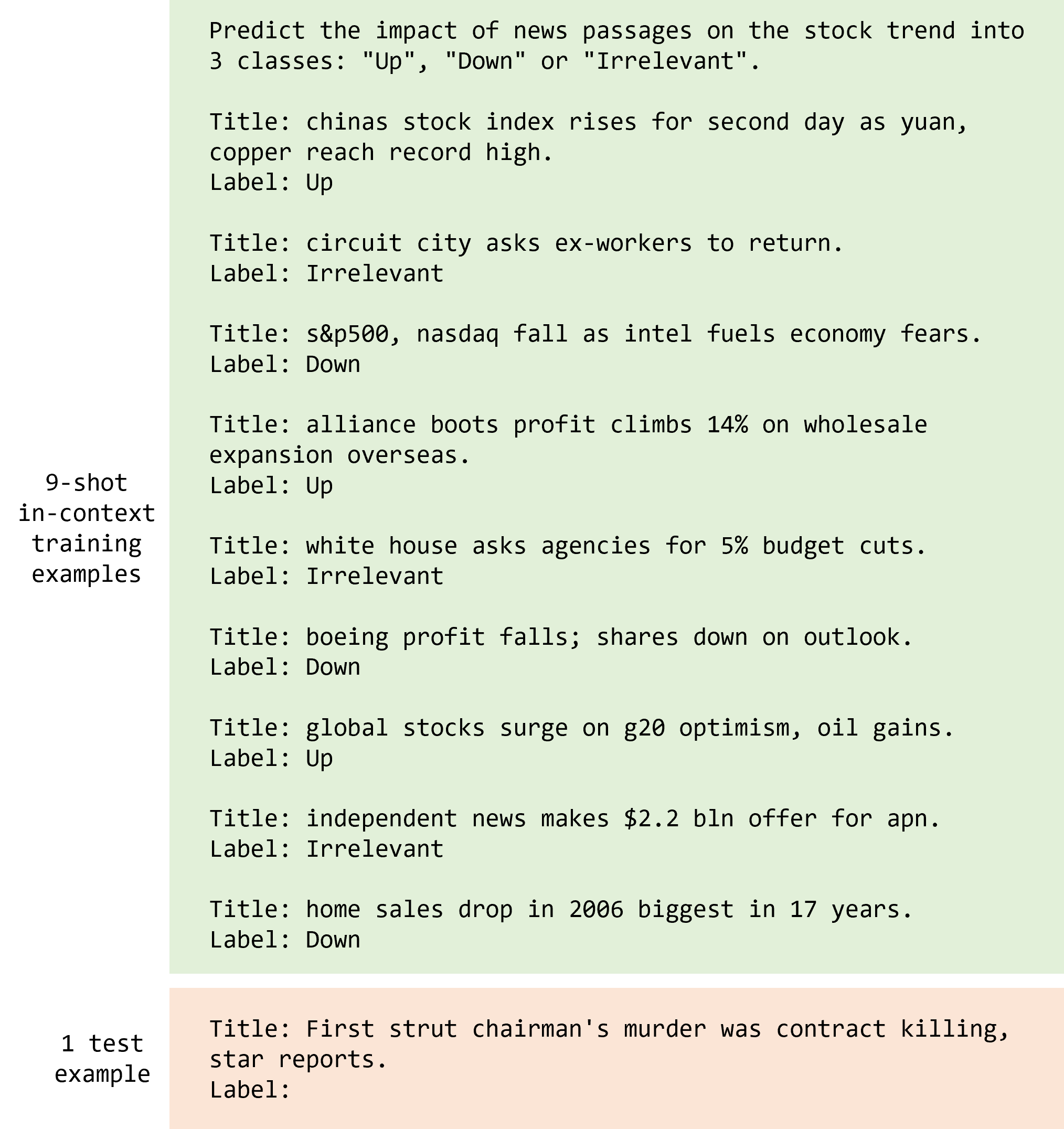}
\caption{The proposed `\textit{denoising-then-voting}' method (9-shots) for S\&P 500 index prediction. All exemplars in the prompt are from different trading days.}
\label{fig:append_sp_3vote}
\end{figure*}

Figure~\ref{fig:append_sp_standard} shows the \textit{standard} prompt we use for U.S. S\&P 500 index prediction. Exemplars in the \textit{standard} prompt are selected from different trading days in the training set. 
`Title' in each exemplar is obtained by concatenating multiple news titles randomly selected within one day. 
`Label' for each exemplar is derived from the corresponding stock movement of that day in the training set.
Figure~\ref{fig:append_sp_2vote} displays the \textit{voting} prompt. 
In the \textit{voting} prompt, `Title' in each exemplar is one news title, which is selected from the corresponding example (the concatenated news titles) in the \textit{standard} prompt. 
We invite five experts with strong financial knowledge to annotate each news title in the \textit{voting} prompt. Figure~\ref{fig:append_sp_3vote} shows our `\textit{denoising-then-voting}' method. 
The `\textit{denoising-then-voting}' method shares the same exemplars of `Up' and `Down' classes with the \textit{voting} prompt. 
In addition, we further provided the exemplars of `Irrelevant'
class. `Title' in each new-added exemplar is also one news title, selected from the other trading days. Still, we ask five financial experts to assess labeling the newly added exemplars. 
As a third category `Irrelevant' being added, we change the beginning instruction of the `\textit{denoising-then-voting}' method as: ``Predict the impact of news passages on the stock trend into 3 classes: ``Up", ``Down" or ``Irrelevant". 

\subsection{Full Few-Shot Prompts for CSI and HK Predictions} \label{append:csi-hk_prompts}
\begin{sidewaysfigure*}[ht]
\centering
\includegraphics[width=\textwidth]{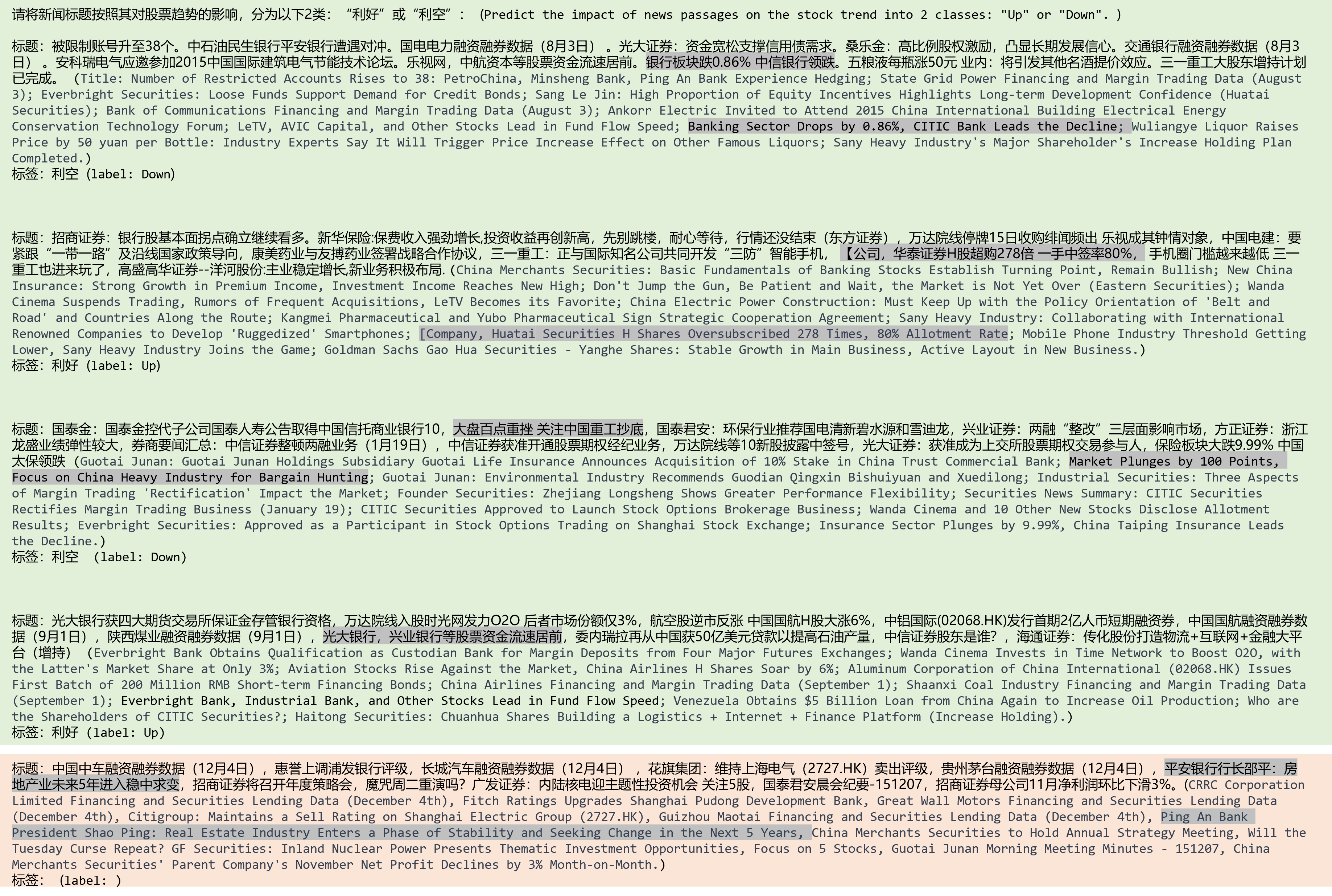}
\caption{The \textit{standard} prompt for CSI-100 \& HK stock trend prediction.}
\label{fig:append_csi_standard}
\end{sidewaysfigure*}

\begin{figure*}[!t]
\centering
\includegraphics[width=0.5\textwidth]{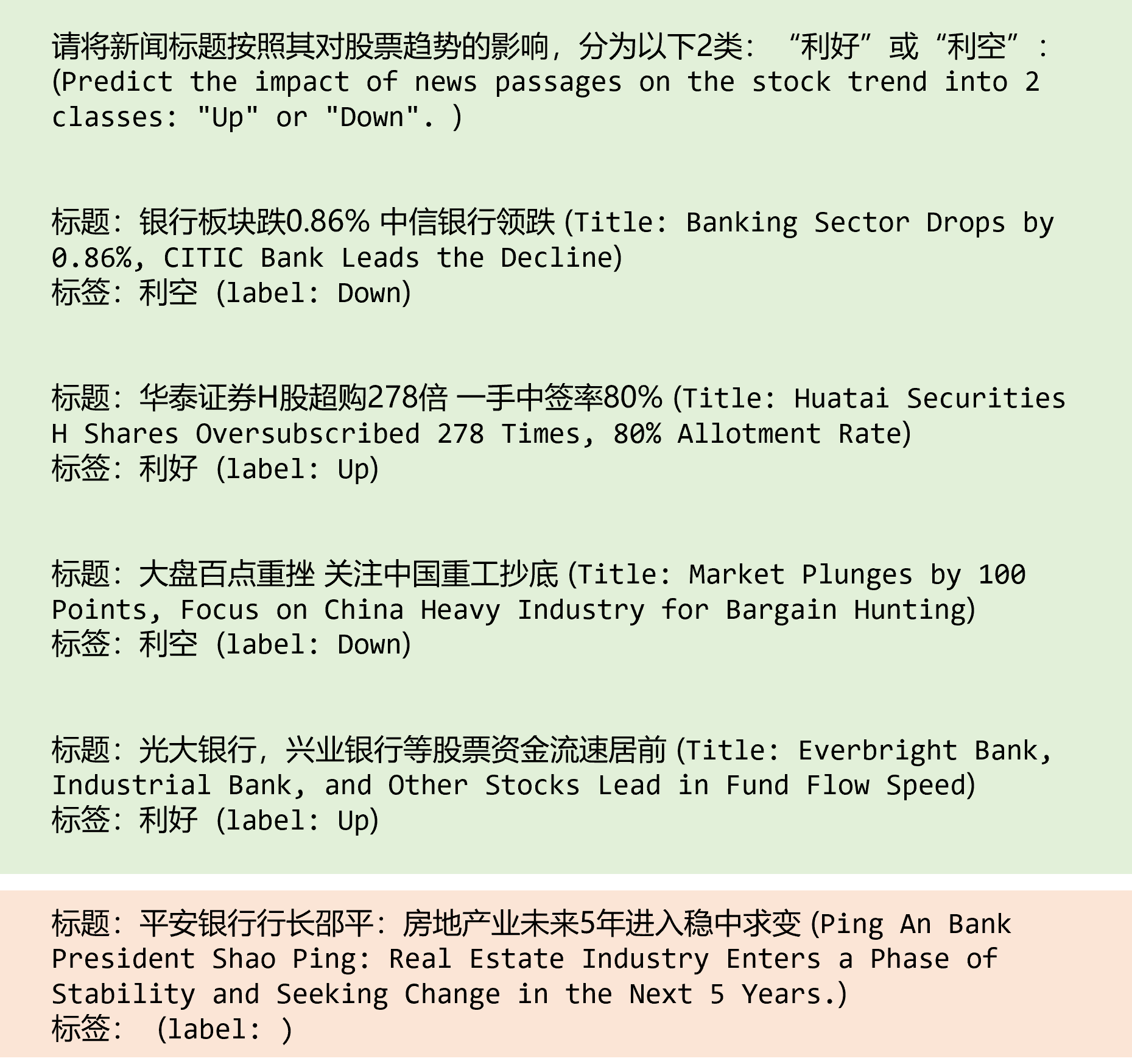}
\caption{The \textit{voting} prompt w/o `Irrelevant' category in CSI-100 \& HK stock trend prediction.}
\label{fig:append_csi-hk_2vote}
\end{figure*}

\begin{figure*}[!t]
\centering
\includegraphics[width=0.65\textwidth]{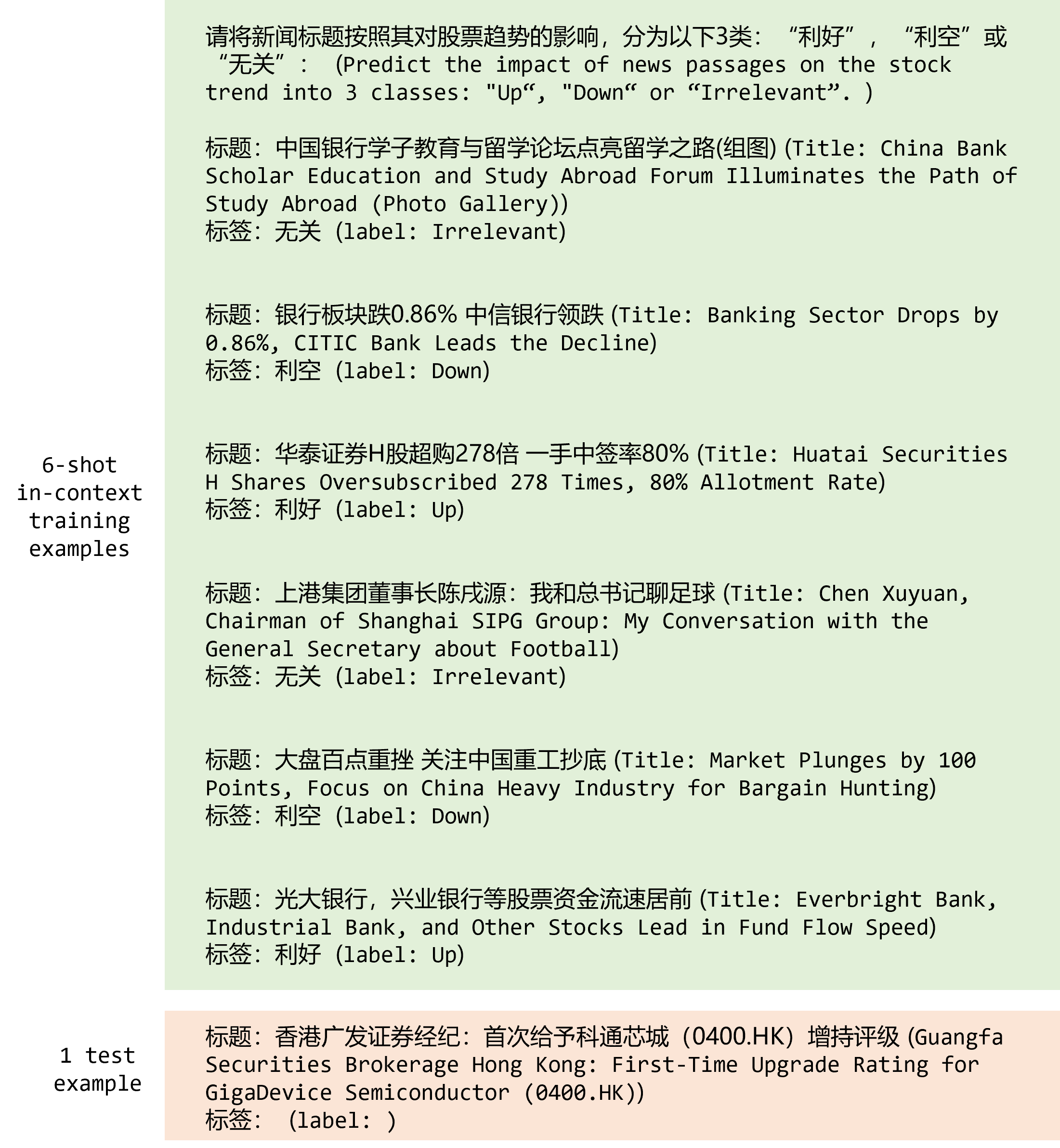}
\caption{The `\textit{denoising-then-voting}' method (6-shots) for CSI-100 \& HK stock trend prediction.}
\label{fig:append_csi-hk_3vote}
\end{figure*}

For both CSI-100 and HK stock prediction, we follow the construction steps 
in \ref{append:sp_prompts} and obtain the \textit{standard}  prompt shown in Figure~\ref{fig:append_csi_standard}, the \textit{voting} prompt in Figure~\ref{fig:append_csi-hk_2vote} and `\textit{denoising-then-voting}' method in Figure~\ref{fig:append_csi-hk_3vote}.

\clearpage

\clearpage

\end{document}